\documentclass{article}
\usepackage{spconf,amsmath,graphicx}
\usepackage[colorlinks=true,
            linkcolor=black,
            citecolor=black,
            urlcolor=black]{hyperref}
\usepackage{booktabs}
\usepackage{multirow}
\usepackage{tabularx}
\usepackage{array}
\usepackage{amssymb}
\usepackage{url}
\usepackage[dvipsnames]{xcolor}

\title{Frequency-guided Multi-level Reasoning for \\ Scene Graph Generation in Video}
\name{Chenxing Li$^{\star \S \dagger}$ \qquad Yiping Duan $^{\star \S \dagger}$ \qquad Xiaoming Tao$^{\star \S \dagger \ddagger}$}
  
  \address{$^{\star}$ Department of Electronic Engineering, Tsinghua University, Beijing, China \\
           $^{\S}$ State Key Laboratory of Space Network and Communications\\
           $^{\dagger}$ Beijing National Research Center for Information Science and Technology (BNRist), Beijing, China\\
           $^{\ddagger}$School of Computer Science and Technology, Xinjiang University, Urumqi, China\thanks{Xiaoming Tao (taoxm@mail.tsinghua.edu.cn) is the corresponding author. This work was supported in part by the National Natural Science Foundation of China  (Nos. NSFC 62322109, 62531012,62595731,  62227801,and U22B2001); in part by the Xplorer Prize in Information and Electronics technologies, and the the Program of Jiangsu Province under Grant NTACT-2024-Z-001.}}

\begin{document}
\ninept
\maketitle
\begin{abstract}
Video Scene Graph Generation aims to obtain structured semantic representations of objects and their relationships in videos for high-level understanding. However, existing methods still have limitations in handling long-tail distributions. This paper proposes the Frequency-guided Relational Multi-level Reasoning (FReMuRe) model, which enhances the modeling ability of long-tail relationships from a mechanism perspective. We introduce relation-specific branches to deal gradient conflicts, yielding more balanced and tail-aware learning. And we design a frequency-aware dual-branch predicate embedding network to model high-frequency and low-frequency relationships separately and improve the recall rate of tail classes through gated fusion. Meanwhile, we propose two types of interchangeable relation classification heads: Bayesian Head for uncertainty estimation and new Gaussian Mixture Model Head to enhance intra-class diversity. Experimental results show that FReMuRe significantly improves the recall rate of long-tail relationships and overall reasoning robustness on the Action Genome dataset. The source code is available at \url{https://github.com/lcx529955/FReMuRe}.
\end{abstract}
\begin{keywords}
Scene Graph Generation, Video representation, Video processing, Long-tail, Frequency-guided
\end{keywords}
\section{Introduction}
\label{sec:intro}

Scene Graph Generation (SGG) translates visual information into a structured format by representing objects as nodes and their relationships as edges. Its application has evolved from static images \cite{wang2025llava, hong2025less, chen2025semantic, li2024object} to video, where it captures not only spatial arrangements, but also crucial dynamic interactions. Video SGG is therefore invaluable for advanced applications like video understanding and event reasoning, as it efficiently extracts key relational semantics from visually dense and often redundant content. However, applying scene graph generation to video faces challenges. A primary issue is temporal incoherence, where processing frames independently causes inconsistencies. The long-tail distribution of predicates also poses a problem, biasing models toward frequent relationships. Furthermore, real-world factors such as occlusions, motion blur, and noisy annotations increase the uncertainty of reasoning.

\begin{figure}[!t]
\centering
\includegraphics[width=8.5cm]{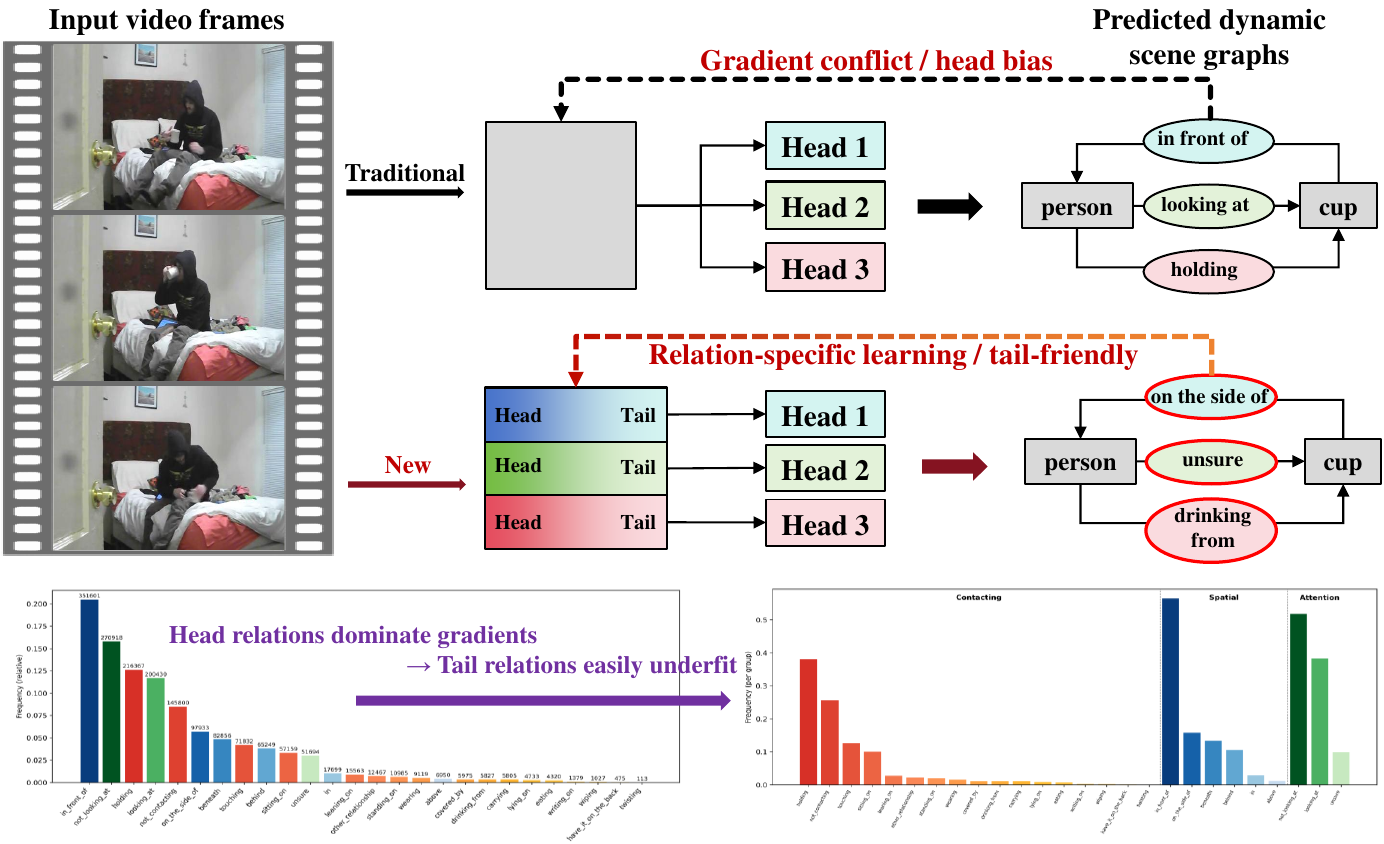}
\caption{\textbf{Motivation of our method.} In Action Genome\cite{ji2020action}, a single shared backbone causes gradient conflicts and head bias under long-tailed distributions. Our relation-specific branches alleviate this, yielding more balanced and tail-aware learning.}
\label{Fig.1}
\end{figure}

\begin{figure*}[!t]
\centering
\includegraphics[width=17cm]{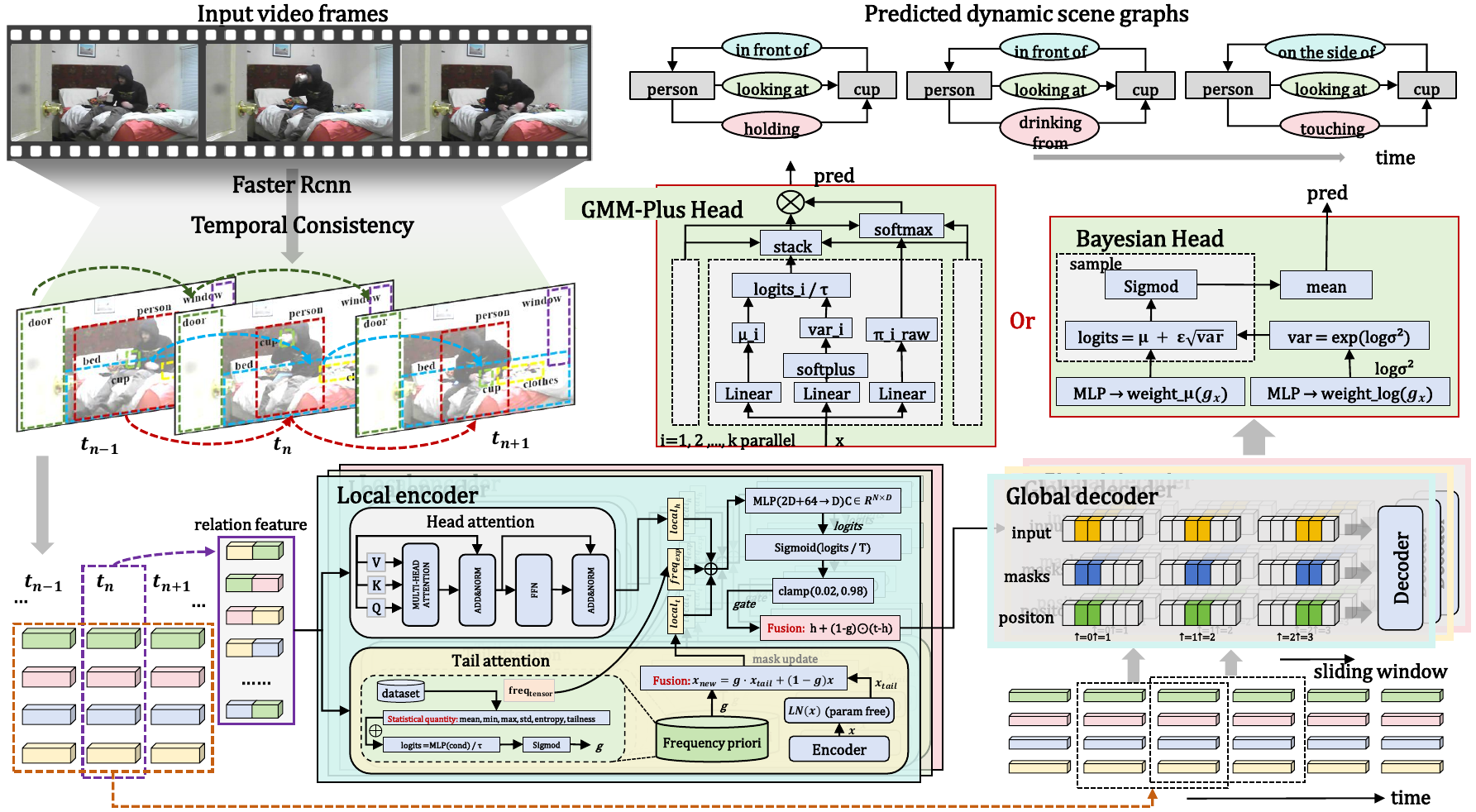}
\caption{\textbf{The overall framework proposed in this paper}. For each video, objects are detected and temporal aligned. Local and global modules model intra-frame and cross-frame relations with three-way decoupling for attention, spatial and contact, while classification is done by GMM-Plus or Bayesian heads. The design supports balanced relation reasoning, long-tail robustness, and uncertainty estimation.}
\label{Fig.2}
\end{figure*}

In response to these challenges, significant progress has been made in Dynamic Scene Graph Generation (DSGG) in recent years \cite{li2022dynamic,gao2022classification,yang20234d,nguyen2024cyclo,rodin2024action}. Existing work such as Temporal Prior Inference (TPI)\cite{wang2022dynamic} and DSG-DETR\cite{feng2023exploiting} has introduced temporal priors and long-term trajectory modeling, enhancing the consistency and dynamic coherence of relationship reasoning. However, these methods mainly focus on temporal modeling and still lack the ability to represent rare relationships. To alleviate the problem of category imbalance, methods such as TEMPURA\cite{nag2023unbiased}, FloCoDe\cite{khandelwal2024flocode}, and TD²-Net\cite{lin2024td2} have improved long-tail category learning to some extent through uncertainty guidance, flow-aware feature debiasing, context denoising, and reweighting strategies. Nevertheless, most of these methods rely on resampling or loss adjustment and have not systematically enhanced the representation ability of tail classes from the relationship modeling mechanism itself. In terms of system process optimization, OED\cite{wang2024oed} simplifies the dynamic graph generation pipeline through a one-stage set prediction architecture, while VirtualHome Action Genome\cite{qiu2023virtualhome} provides better data support for subsequent modeling by improving the density of data annotations and extending long-term semantic information. However, these methods mainly focus on overall process and data optimization and have not deep-recognized fine-grained modeling of tail-class relationships. Furthermore, studies such as Situation HyperGraphs\cite{urooj2023learning} and HiLo\cite{zhou2023hilo} have attempted to enhance reasoning capabilities from the perspectives of structural generalization and frequency relationship modeling, proposing new strategies such as hyperedge modeling and joint debiasing of high and low frequencies. 

Although these provide inspiration for complex relationship modeling, they are mostly concentrated in static images or general reasoning tasks and lack specialized optimization for efficient semantic encoding of long-tail relationships in video scene graphs. As shown in Fig.\ref{Fig.1}, different types of relationships share the same backbone network, leading to gradient conflicts and head-class dominance bias, making it difficult to effectively learn rare relationships. To address this, we propose the Frequency-guided Relational Multi-level Reasoning (FReMuRe) model. Instead of forcing diverse relationship types through a unified feature space, our model decouples the learning process based on predicate frequency. This allows for dedicated modeling of both common and rare relationships, fundamentally resolving the issue of gradient conflict. Meanwhile, it also innovatively combines frequency-aware mechanisms, dual-branch predicate embedding modeling, and Bayesian/ GMM-Plus classification head optimization to systematically enhance the robustness of long-tail relationship representation and reasoning. The main contributions of this paper are as follows.

\noindent 1. \textbf{Frequency-guided tail-class debiasing}: A frequency-aware gating mechanism is employed to model tail-class relationships, and prior knowledge of frequencies was introduced to improve the recall of low-frequency relationships.

\noindent 2. \textbf{Decoupled Dual-Branch Network for Predicate Embedding}: We introduce a dual-branch Transformer to decouple the learning process for high- and low-frequency relationships. This architectural separation is designed to resolve gradient conflicts, ensuring a dynamic balance in modeling both head and tail classes.

\noindent 3. \textbf{Replaceable relationship classification heads}: Two options are provided: the BayesianHead models uncertainty for robust predictions, while GMM-Plus improves tail-class representation by incorporating multiple prototypes and adjustable variances.

\begin{table*}[!t]
\centering
\caption{Comparative results for DSGG tasks: PREDCLS and SGCLS on AG, \textbf{Bold} is the best result, and \underline{underline} is the second-best
\label{table 1}}
\resizebox{\linewidth}{!}{
\begin{tabular}{l|ccc|ccc|ccc}
\toprule
\multirow{2}{*}{Method} &
\multicolumn{3}{c|}{predcls} &
\multicolumn{3}{c|}{sgcls} &
\multicolumn{3}{c}{sgdet} \\
\cline{2-4} \cline{5-7} \cline{8-10}

& R/mR@10 & R/mR@20 & R/mR@50 & R/mR@10 & R/mR@20 & R/mR@50 & R/mR@10 & R/mR@20 & R/mR@50 \\

\toprule

RelDN\cite{zhang2019graphical}
& 20.3 / 6.2 & 20.3 / 6.2 & 20.3 / 6.2 & 11.0 / 3.4 & 11.0 / 3.4 & 11.0 / 3.4 & 9.1 / 3.3 & 9.1 / 3.3 & 9.1 / 3.3 \\

TRACE\cite{teng2021target}
& 27.5 / 15.2 & 27.5 / 15.2 & 27.5 / 15.2 & 14.8 / 8.9 & 14.8 / 8.9 & 14.8 / 8.9 & 13.9 / 8.2 & 14.5 / 8.2 & 14.5 / 8.2 \\

STTran\cite{cong2021spatial}
& 68.6 / 37.8 & 71.8 / 40.1 & 71.8 / 40.2 & 45.3 / 27.2 & 46.0 / 28.0 & 46.1 / 28.0 & \textbf{25.8} / \textbf{16.6} & 33.0 / \textbf{20.8} & 33.3 / 22.2 \\

STTran\text{-}TPI\cite{wang2022dynamic}
& \textbf{69.7} / 37.3 & 71.6 / 40.6 & 71.6 / 40.6 & 45.3 / 28.3 & 46.1 / 29.3 & 46.1 / 29.3 & 25.1 / \underline{15.6} & 32.9 / 20.2 & 33.2 / 21.8 \\

TEMPURA\cite{nag2023unbiased}
& 68.5 / 37.5 & 71.1 / 40.8 & 71.2 / 40.9 & 45.3 / 28.4 & \underline{46.2} / 29.5 & 46.3 / 29.5 & 25.3 / 15.5 & \underline{33.1} / 20.4 & \underline{35.2} / \underline{21.9} \\

\hline

FReMuRe+bayesian(ours)
& \underline{69.5} / \textbf{38.9} & \textbf{72.5} / \underline{42.9} & \textbf{72.5} / \textbf{43.1}
& \underline{45.4} / \textbf{29.4} & 45.5 / \underline{30.5} & \underline{46.5} / \textbf{30.6}
& 25.4 / 14.9 & \underline{33.1} / 20.1 & \underline{35.2} / \underline{21.9} \\

FReMuRe+gmm-plus(ours)
& 69.1 / \underline{38.7} & \underline{72.2} / \textbf{43.1} & \underline{72.2} / \textbf{43.1}
& \textbf{45.5} / \textbf{29.4} & \textbf{46.6} / \textbf{30.6} & \textbf{46.6} / \textbf{30.6}
& \underline{25.6} / 15.1 & \textbf{33.6} / \underline{20.7} & \textbf{36.1} / \textbf{22.5} \\
\bottomrule
\end{tabular}
}
\end{table*}

\section{Method}
\label{sec:format}

\subsection{Framework Overview}

We propose the \textbf{FReMuRe} model, as shown in Fig.\ref{Fig.2}. FReMuRe first detects objects using Faster R-CNN and temporal consistency module \cite{nag2023unbiased}. A \textbf{dual-branch predicate network} then processes object pairs on decoupled pathways to model relationships, implementing our core strategy. The final predictions are generated by a global decoder and an enhanced Bayesian head or GMM-Plus head.

Additionally, our work addresses the \textbf{negative transfer} caused by gradient conflicts in models that utilize a shared feature extractor ($\theta$) for diverse relation types. In such a paradigm, the aggregated gradient, $\nabla_{\theta} L = \nabla_{\theta} L_a + \nabla_{\theta} L_s + \nabla_{\theta} L_c$, can contain conflicting components (i.e., $\langle \nabla_{\theta} L_a, \nabla_{\theta} L_s \rangle < 0$), which impairs the model's ability to learn representations for long-tail relations. To resolve this, our central principle is to decompose the learning process. By dedicating independent parameters to each relation type, we enable disentangled updates, $\nabla_{\theta_a} L_a, \nabla_{\theta_s} L_s, \nabla_{\theta_c} L_c$, allowing each to learn a specialized representation without destructive interference.

The following will provide a detailed introduction to the specifics of each module.

\subsection{Frequency-guided mechanism}

Relation prediction in video scene graphs intrinsically follows a severe long-tail distribution: frequent relations (e.g., ``holding'') dominate the dataset, while rare ones (e.g., ``carrying'') are significantly underrepresented. To address this, we propose a Frequency-guided mechanism that exploits category frequency—strictly referring to \textbf{statistical occurrence} in this work—to adaptively recalibrate relation features and strengthen tail-class modeling.

\textbf{Relationship frequency modeling.} Let the global relation set be denoted as $R = \{r_1, r_2, ..., r_{|R|}\}$. We compute the category frequency $f_k = n_k/\sum_{j=1}^{|R|} n_j$, where $n_k$ represents the total occurrence count of relation $r_k$ and $\sum n_j$ is the total count. We then construct a frequency tensor $f = [f_1, f_2, ..., f_{|R|}] \in \mathbb{R}^{|R|}$, which serves as the prior information for subsequent bias correction.

\textbf{Frequency-aware correction.} We scale the input relation features $x \in \mathbb{R}^d$ using an inverse-frequency weighting scheme. The correction process is formulated as:

\vspace{-5pt}
\begin{equation}\label{eq1}
x' = g(f) \odot LN(x) + \big(1 - g(f)\big) \odot x
\end{equation}
\vspace{-10pt}

\noindent where $LN(\cdot)$ is a parameter-free LayerNorm that stabilizes the feature distribution of tail categories, and $\odot$ denotes element-wise multiplication. The term $g(f)$ is a learnable frequency gate defined as:

\vspace{-5pt}
\begin{equation}\label{eq2}
g(f) = \sigma\big(W \cdot \log(1/f + \epsilon) + b\big)
\end{equation}
\vspace{-10pt}

\noindent where $\sigma$ is the sigmoid function ensuring bounded outputs, $b$ is the bias term, and $\epsilon$ is a smoothing term. Crucially, the term $\log(1/f)$ mathematically converts frequency into a feature scaling factor based on \textbf{self-information theory}. This design strictly \textbf{counteracts the gradient dominance} of head classes by enlarging gate values for rare categories. Unlike simple loss-level reweighting, this mechanism constitutes a structural correction, improving the recognition and robustness of tail relations within the representation itself.

\subsection{Dual-branch Predicate Embedding Generator}

To effectively balance the learning of high-frequency (head) and low-frequency (tail) relations, we propose the Dual-branch Predicate Embedding Generator (DPEG). Given the input sequence of object pairs $(o^t_i; o^t_j)$ from each frame, an initial relationship embedding $x_{ij}^t \in \mathbb{R}^d$ is first generated and then processed through specific transformations before entering the DPEG.

\vspace{-5pt}
\begin{equation}\label{eq3}
G_{loc} = \sigma \left(W_{loc} \left[ H_{loc} \| T_{loc} \| f \right] + b_{loc} \right)
\end{equation}
\vspace{-10pt}

\vspace{-5pt}
\begin{equation}\label{eq4}
Z_{loc} = G_{loc} \odot H_{loc} + (1 - G_{loc}) \odot T_{loc}
\end{equation}
\vspace{-10pt}

In the local branch phase, DPEG computes two parallel branches. The standard local encoding path outputs $H_{loc}$ as the feature representation for high-frequency relations. Simultaneously, the frequency-aware Tail branch uses $T_{loc}$ to specifically model low-frequency relations. It is worth noting that this dual-branch architecture and the gating mechanism are \textbf{complementary rather than redundant}: the former physically decouples parameters to resolve gradient conflicts, while the latter handles dynamic feature as shown in Function (\ref{eq3}) and (\ref{eq4}), here $\|$ indicates concatenation, and training is stabilized via LayerNorm and zero-bias initialization for the gate to prevent early saturation.

In the global branch phase, we incorporate cross-frame information by introducing positional encoding to complement the global context of the target relations. Specifically, the global branch processes the sequence to produce the final relation classification needed for the global representation $Z_{glob}$. Finally, we pass $Z_{glob}$ through a sliding window mechanism. If the same relationship appears within multiple overlapping windows, we perform \textbf{average or weighted aggregation} to obtain the final refined embedding $E_t \in \mathbb{R}^{L \times d}$. This aggregation strategy is vital for refining prediction consistency across temporal contexts before the memory diffusion modules.

\subsection{Bayesian and GMM-Plus Classification Heads}

In dynamic scene graph generation, predicate classifiers struggle with long-tail distributions and annotation noise. FReMuRe introduces the Bayesian and GMM-Plus Heads to address these issues.

\textbf{Bayesian Head.} This head models the prediction as a distribution. It predicts the mean $\mu$ and variance $\sigma^2$ for each relation class, with variance constrained by $\sigma^2 = \exp(\log(\sigma^2))$. During training, we employ Monte Carlo sampling:

\vspace{-5pt}
\begin{equation}\label{eq5}
\hat{z} = \mu + \epsilon \cdot \sqrt{\sigma^2}, \quad \epsilon \sim \mathcal{N}(0,1)
\end{equation}
\vspace{-10pt}

\noindent Multiple samples are drawn to estimate the class distribution, and the output is the mean probability prediction. This probabilistic formulation not only improves tail robustness but also captures two types of uncertainty: \textit{aleatoric uncertainty} arising from inherent data variance, and \textit{epistemic uncertainty} estimated from the entropy of the predicted distribution.

\textbf{GMM-Plus Head.} An extension of the traditional Gaussian Mixture Model (GMM), each relation class is composed of $K$ Gaussian components $(\mu_k, \sigma_k^2, \pi_k)$. The output probability is:

\vspace{-5pt}
\begin{equation}\label{eq6}
p(y|z) = \sum_{k=1}^{K} \pi_k \cdot \mathcal{N}(z|\mu_k, \sigma_k^2)
\end{equation}
\vspace{-10pt}

\noindent where $\pi_k$ is the mixing coefficient normalized across classes using softmax. During training, the component mean $\mu_k$ is updated with Gaussian sampling perturbation, while the variance $\sigma_k^2$ is kept stable using softplus and clamping. Unlike traditional GMM \cite{choi2021active}, GMM-Plus introduces a frequency-aware regularization term to capture \textbf{diverse visual modes}. Crucially, this constraint prevents Gaussian components from degenerating into \textbf{Dirac distributions} (i.e., $\sigma^2 \to 0$) when tail samples are scarce. By enforcing a lower bound on variance, we ensure robust multi-modal modeling and prevent overfitting to few-shot examples.


\section{Experiments}

\subsection{Experiments setting}

\textbf{Dataset.} We conduct experiments on the Action Genome (AG) \cite{ji2020action} dataset, which provides dense dynamic scene graph annotations for benchmarking. It includes 35 objects except people and 25 types of relationships in attention, spatial and contact. The videos in dataset were filtered and sorted into 7584 train sets and 1750 test sets. 

\noindent \textbf{Evaluation Setup.} We employ distinct loss functions for different relation types: cross-entropy for single-label attention relations, and binary cross-entropy or multi-label margin loss for multi-label spatial and contact relations. The model is trained for 10 epochs using the Adam \cite{2014Adam} optimizer.

\noindent \textbf{Metrics.} We use two standard metrics \cite{41misra2016seeing}. Recall@K (R@K) measures the fraction of true relationships captured within the top-K highest-confidence predictions. To better assess performance on long-tailed distributions, we also use Mean Recall@K (mR@K), which calculates the average R@K across all relationship categories.

\subsection{Results and Discussions}
 
We evaluated our model on three SGG tasks, namely: (1) Predicate classification (PREDCLS): Prediction of predicate labels of object pairs, given ground truth labels and bounding boxes of objects; (2) Scene graph classification (SGCLS): Joint classification of predicate labels and the ground truth bounding boxes; (3) Scene graph detection (SGDET): End-to-end detection of the objects and predicate classification of object pairs.

\textbf{Quantitative Results:} As shown in Table \ref{table 1}, our FReMuRe variants achieve state-of-the-art performance: RelDN \cite{zhang2019graphical}, TRACE \cite{teng2021target}, STTran \cite{cong2021spatial}, STTran \text{-}TPI \cite{wang2022dynamic} and TEMPURA \cite{nag2023unbiased}, especially on the mean Recall@K (mR@K) metric, which is crucial for evaluating long-tail performance. On the PREDCLS task, our model shows particular strength in mean recall. FReMuRe+gmm-plus achieves the top score for mR@50 (43.1), surpassing prior methods. This result highlights its effective debiasing capability. On the SGCLS task, FReMuRe+gmm-plus sets new state-of-the-art results across all reported metrics, reaching 46.6 for R@50 and 30.6 for mR@50, which indicates its robustness. On the challenging SGDET task, while some models perform well at lower K values, FReMuRe+gmm-plus obtains the best scores for R@50 (36.1) and mR@50 (22.5). This suggests our model more effectively captures a diverse set of relationships, including less frequent ones, rather than focusing only on the most common predictions.

\begin{figure}[!t]
\centering
\includegraphics[width=8.5cm]{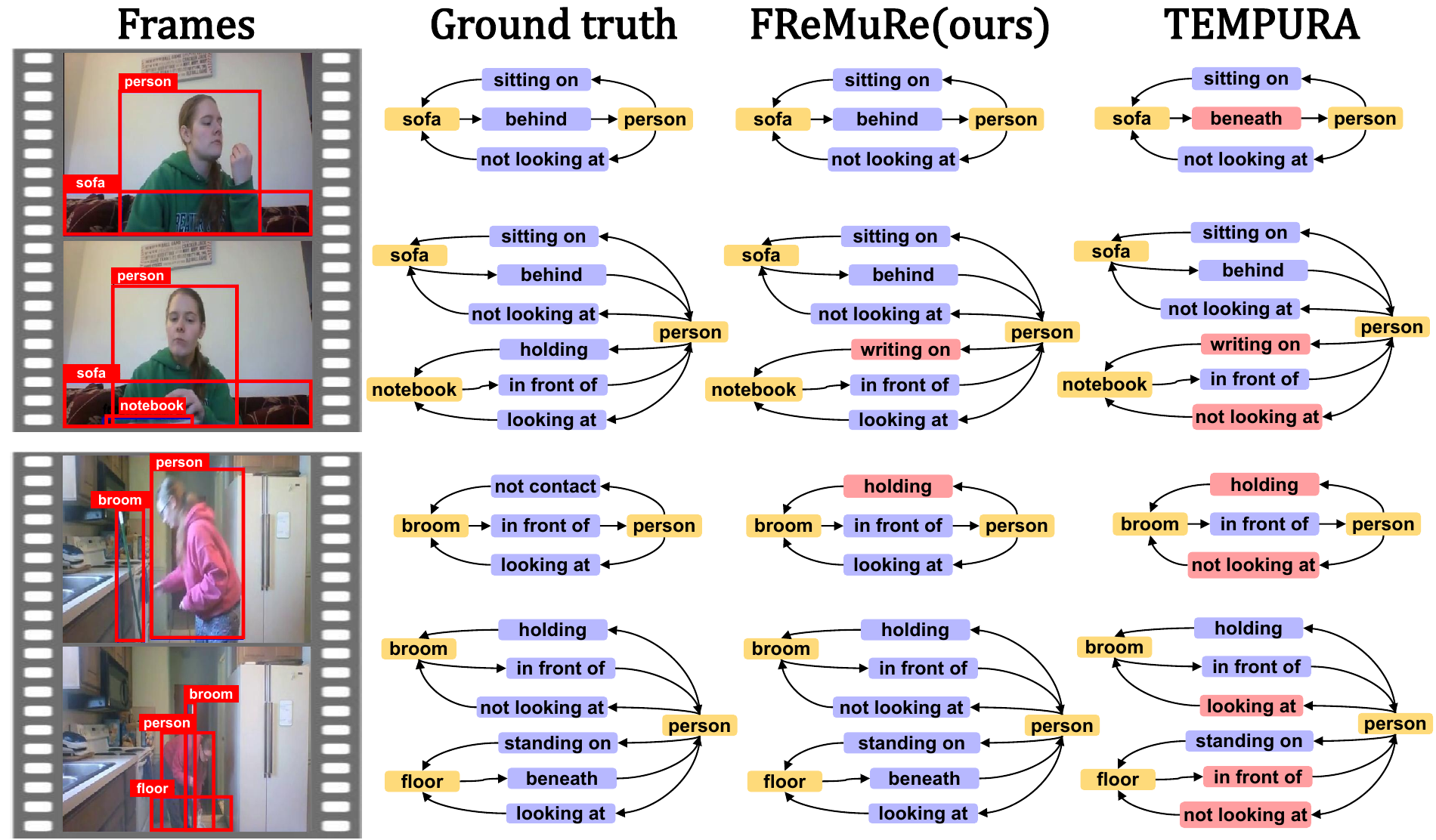}
\caption{\textbf{Comparative qualitative results}. From left to right: input video frames, ground truth scene graphs, scene graphs generated by FReMuRe. Incorrect predicate predictions are shown in pink}
\label{Fig.3}
\end{figure}

\textbf{Qualitative Results:} As illustrated in Fig.\ref{Fig.3}, FReMuRe generates scene graphs that more closely align with the ground truth compared to baselines like TEMPURA. For example, our model correctly identifies the ``person looking at notebook" relationship, which TEMPURA misclassifies as ``not looking at". Similarly, in the bottom scene, FReMuRe accurately detects that the person is ``standing on" the floor, whereas TEMPURA incorrectly predicts ``in front of". This improved accuracy on less frequent interactions can be attributed to our decoupled learning mechanism, which mitigates the inherent bias toward common relationships.

\subsection{Ablation Studies}
\label{sec:subhead}

To validate the contribution of each component in our FReMuRe framework, we conducted ablation studies with results shown in Table \ref{table 2}. These experiments confirm that our overall design and each of its modules are integral to performance, as removing any single part leads to a notable degradation in mR@K.

\textbf{Core Decoupling Strategy:} The central principle of decoupling the learning process is the most critical factor. The no decouple variant suffers the most severe performance degradation across all tasks, with the SGDET mR@50 score plummeting from 22.5 to 17.3. This provides strong evidence that mitigating gradient conflicts through separate learning pathways is essential. \textbf{Architectural Components:} The specific mechanisms implementing our strategy are also vital. Removing the Dual-Branch Architecture (no dual-branch) or the Frequency-Guided Mechanism (no frequency) also causes a drastic drop in performance, with SGCLS mR@10 falling to 24.3 and 25.1 respectively. \textbf{Specialized Classification Heads: }Finally, the specialized heads provide a significant performance boost. Removing the GMM-Plus head (no gmm-plus) lowers the SGDET mR@50 score from 22.5 to 19.4, while ablating the Bayesian head (no bayes) shows a similar impact, validating their respective roles in handling complex and uncertain relations.

\begin{table}[t]
\centering
\caption{Ablation on Action Genome (mR@K)}
\label{table 2}
\setlength{\tabcolsep}{3pt} 
\resizebox{\linewidth}{!}{
\begin{tabular}{l|c|c|c}
\toprule
\multirow{2}{*}{Module} &
\multicolumn{1}{c|}{predcls} &
\multicolumn{1}{c|}{sgcls} &
\multicolumn{1}{c}{sgdet} \\
\cline{2-4} 
& mR@10/20/50 & mR@10/20/50 & mR@10/20/50 \\
\toprule
no decouple     & 36.3 / 40.4 / 41.4   & 24.8 / 28.5 / 28.6 & 12.1 / 17.3 / 17.3 \\
no frequency    & 36.8 / 40.9 / 41.0   & 25.1 / 29.8 / 30.2 & 12.5 / 18.1 / 19.7 \\
no dual-branch  & 36.5 / 40.1 / 40.1   & 24.3 / 28.7 / 29.2 & 12.5 / 18.8 / 19.9 \\
no bayes        & 37.5 / 41.2 / 41.3   & 26.5 / 30.1 / 30.6 & 13.4 / 19.6 / 20.3 \\
no gmm-plus     & 36.6 / 40.7 / 40.2   & 24.4 / 29.5 / 29.9 & 12.8 / 18.7 / 19.4 \\
\hline
\textit{\textbf{ours+bayesian}} & \textbf{38.9} / \textbf{42.9} / \textbf{43.1} & \textbf{29.4} / \textbf{30.5} / \textbf{30.6} & \textbf{14.9} / \textbf{20.1} / \textbf{21.9 }\\
\textit{\textbf{ours+gmm-plus}} & \textbf{38.7} / \textbf{43.1} / \textbf{43.1} & \textbf{29.4} / \textbf{30.6} / \textbf{30.6} & \textbf{15.1} / \textbf{20.7} / \textbf{22.5} \\
\bottomrule
\end{tabular}
}
\end{table}

\section{Conclusions}
\label{sec:typestyle}

This paper proposed FReMuRe model that enhances the representation of rare relationships in video scene graph generation. By employing a frequency-guided dual-branch network and specialized classification heads, FReMuRe effectively decouples the learning of common and rare predicates. Our experiments on the Action Genome dataset confirm the model's success, showing clear improvements on tail-class prediction and achieving state-of-the-art mR@K scores. Ablation studies further validate that our decoupled design is crucial to this performance. Despite these advances, future work will focus on tackling challenges posed by extremely imbalanced data and adapting the model to new domains.

\bibliographystyle{IEEEbib}
\bibliography{refs}

@inproceedings{nag2023unbiased,
  title={Unbiased Scene Graph Generation in Videos},
  author={Nag, Sayak and Min, Kyle and Tripathi, Subarna and Roy-Chowdhury, Amit K},
  booktitle={Proceedings of the IEEE/CVF Conference on Computer Vision and Pattern Recognition},
  pages={22803--22813},
  year={2023}
}

@inproceedings{wang2022dynamic,
  title={Dynamic scene graph generation via temporal prior inference},
  author={Wang, Shuang and Gao, Lianli and Lyu, Xinyu and Guo, Yuyu and Zeng, Pengpeng and Song, Jingkuan},
  booktitle={Proceedings of the 30th ACM International Conference on Multimedia},
  pages={5793--5801},
  year={2022}
}

@inproceedings{khandelwal2024flocode,
  title={Flocode: Unbiased dynamic scene graph generation with temporal consistency and correlation debiasing},
  author={Khandelwal, Anant},
  booktitle={Proceedings of the IEEE/CVF Conference on Computer Vision and Pattern Recognition},
  pages={2516--2526},
  year={2024}
}

@inproceedings{lin2024td2,
  title={Td$^2$-net: Toward denoising and debiasing for video scene graph generation},
  author={Lin, Xin and Shi, Chong and Zhan, Yibing and Yang, Zuopeng and Wu, Yaqi and Tao, Dacheng},
  booktitle={Proceedings of the AAAI Conference on Artificial Intelligence},
  volume={38},
  pages={3495--3503},
  year={2024}
}

@inproceedings{qiu2023virtualhome,
  title={Virtualhome action genome: A simulated spatio-temporal scene graph dataset with consistent relationship labels},
  author={Qiu, Yue and Nagasaki, Yoshiki and Hara, Kensho and Kataoka, Hirokatsu and Suzuki, Ryota and Iwata, Kenji and Satoh, Yutaka},
  booktitle={Proceedings of the IEEE/CVF Winter Conference on Applications of Computer Vision},
  pages={3351--3360},
  year={2023}
}

@inproceedings{zhou2023hilo,
  title={Hilo: Exploiting high low frequency relations for unbiased panoptic scene graph generation},
  author={Zhou, Zijian and Shi, Miaojing and Caesar, Holger},
  booktitle={Proceedings of the IEEE/CVF international conference on computer vision},
  pages={21637--21648},
  year={2023}
}

@inproceedings{wang2024oed,
  title={Oed: Towards one-stage end-to-end dynamic scene graph generation},
  author={Wang, Guan and Li, Zhimin and Chen, Qingchao and Liu, Yang},
  booktitle={Proceedings of the IEEE/CVF conference on computer vision and pattern recognition},
  pages={27938--27947},
  year={2024}
}

@inproceedings{urooj2023learning,
  title={Learning situation hyper-graphs for video question answering},
  author={Urooj, Aisha and Kuehne, Hilde and Wu, Bo and Chheu, Kim and Bousselham, Walid and Gan, Chuang and Lobo, Niels and Shah, Mubarak},
  booktitle={Proceedings of the IEEE/CVF Conference on Computer Vision and Pattern Recognition},
  pages={14879--14889},
  year={2023}
}

@inproceedings{feng2023exploiting,
  title={Exploiting long-term dependencies for generating dynamic scene graphs},
  author={Feng, Shengyu and Mostafa, Hesham and Nassar, Marcel and Majumdar, Somdeb and Tripathi, Subarna},
  booktitle={Proceedings of the IEEE/CVF winter conference on applications of computer vision},
  pages={5130--5139},
  year={2023}
}

@inproceedings{ji2020action,
  title={Action genome: Actions as compositions of spatio-temporal scene graphs},
  author={Ji, Jingwei and Krishna, Ranjay and Fei-Fei, Li and Niebles, Juan Carlos},
  booktitle={Proceedings of the IEEE/CVF conference on computer vision and pattern recognition},
  pages={10236--10247},
  year={2020}
}

@article{li2024object,
  title={Object--attribute--relation model-based semantic coding for image transmission},
  author={Li, Chenxing and Duan, Yiping and Tao, Xiaoming and Hu, Shuzhan and Yang, Qianqian and Chen, Changwen},
  journal={Journal of the Franklin Institute},
  volume={361},
  number={11},
  pages={106942},
  year={2024},
  publisher={Elsevier}
}

@inproceedings{wang2025llava,
  title={LLaVA-SG: Leveraging scene graphs as visual semantic expression in vision-language models},
  author={Wang, Jingyi and Ju, Jianzhong and Luan, Jian and Deng, Zhidong},
  booktitle={ICASSP 2025-2025 IEEE International Conference on Acoustics, Speech and Signal Processing (ICASSP)},
  pages={1--5},
  year={2025},
  organization={IEEE}
}

@inproceedings{hong2025less,
  title={Less is more: Efficient Scene Graph Generation with reparameterization},
  author={Hong, Jonghwan and Noh, Seonghyeok and Ku, Bonhwa and Ko, Hanseok},
  booktitle={ICASSP 2025-2025 IEEE International Conference on Acoustics, Speech and Signal Processing (ICASSP)},
  pages={1--5},
  year={2025},
  organization={IEEE}
}

@inproceedings{chen2025semantic,
  title={Semantic Graph Embedded Energy Minimization Learning for Scene Graph Generation},
  author={Chen, Jinghang and Zhang, Chi and Liu, Yuehu and Wang, Le},
  booktitle={ICASSP 2025-2025 IEEE International Conference on Acoustics, Speech and Signal Processing (ICASSP)},
  pages={1--5},
  year={2025},
  organization={IEEE}
}

@inproceedings{li2022dynamic,
  title={Dynamic scene graph generation via anticipatory pre-training},
  author={Li, Yiming and Yang, Xiaoshan and Xu, Changsheng},
  booktitle={Proceedings of the IEEE/CVF conference on computer vision and pattern recognition},
  pages={13874--13883},
  year={2022}
}

@inproceedings{gao2022classification,
  title={Classification-then-grounding: Reformulating video scene graphs as temporal bipartite graphs},
  author={Gao, Kaifeng and Chen, Long and Niu, Yulei and Shao, Jian and Xiao, Jun},
  booktitle={Proceedings of the IEEE/CVF conference on computer vision and pattern recognition},
  pages={19497--19506},
  year={2022}
}

@article{yang20234d,
  title={4d panoptic scene graph generation},
  author={Yang, Jingkang and Cen, Jun and Peng, Wenxuan and Liu, Shuai and Hong, Fangzhou and Li, Xiangtai and Zhou, Kaiyang and Chen, Qifeng and Liu, Ziwei},
  journal={Advances in Neural Information Processing Systems},
  volume={36},
  pages={69692--69705},
  year={2023}
}

@article{nguyen2024cyclo,
  title={CYCLO: Cyclic graph transformer approach to multi-object relationship modeling in aerial videos},
  author={Nguyen, Trong-Thuan and Nguyen, Pha and Li, Xin and Cothren, Jackson and Yilmaz, Alper and Luu, Khoa},
  journal={Advances in Neural Information Processing Systems},
  volume={37},
  pages={90355--90383},
  year={2024}
}

@inproceedings{rodin2024action,
  title={Action scene graphs for long-form understanding of egocentric videos},
  author={Rodin, Ivan and Furnari, Antonino and Min, Kyle and Tripathi, Subarna and Farinella, Giovanni Maria},
  booktitle={Proceedings of the IEEE/CVF Conference on Computer Vision and Pattern Recognition},
  pages={18622--18632},
  year={2024}
}

@inproceedings{choi2021active,
  title={Active learning for deep object detection via probabilistic modeling},
  author={Choi, Jiwoong and Elezi, Ismail and Lee, Hyuk-Jae and Farabet, Clement and Alvarez, Jose M},
  booktitle={Proceedings of the IEEE/CVF international conference on computer vision},
  pages={10264--10273},
  year={2021}
}

@inproceedings{zhang2019graphical,
  title={Graphical contrastive losses for scene graph parsing},
  author={Zhang, Ji and Shih, Kevin J and Elgammal, Ahmed and Tao, Andrew and Catanzaro, Bryan},
  booktitle={Proceedings of the IEEE/CVF conference on computer vision and pattern recognition},
  pages={11535--11543},
  year={2019}
}

@inproceedings{teng2021target,
  title={Target adaptive context aggregation for video scene graph generation},
  author={Teng, Yao and Wang, Limin and Li, Zhifeng and Wu, Gangshan},
  booktitle={Proceedings of the IEEE/CVF International Conference on Computer Vision},
  pages={13688--13697},
  year={2021}
}

@inproceedings{cong2021spatial,
  title={Spatial-temporal transformer for dynamic scene graph generation},
  author={Cong, Yuren and Liao, Wentong and Ackermann, Hanno and Rosenhahn, Bodo and Yang, Michael Ying},
  booktitle={Proceedings of the IEEE/CVF international conference on computer vision},
  pages={16372--16382},
  year={2021}
}

@article{2014Adam,
  title={Adam: A Method for Stochastic Optimization},
  author={ Kingma, D.  and  Ba, J. },
  journal={Computer Science},
  year={2014},
}

@inproceedings{41misra2016seeing,
  title={Seeing through the human reporting bias: Visual classifiers from noisy human-centric labels},
  author={Misra, Ishan and Lawrence Zitnick, C and Mitchell, Margaret and Girshick, Ross},
  booktitle={Proceedings of the IEEE conference on computer vision and pattern recognition},
  pages={2930--2939},
  year={2016}
}

\end{document}